\documentclass[runningheads]{llncs}

\usepackage[T1]{fontenc}
\usepackage{graphicx}
\usepackage{booktabs}
\usepackage[misc]{ifsym}
\usepackage{amsmath}
\usepackage{amssymb}
\usepackage{xcolor}
\usepackage{rotating}
\usepackage{subcaption}
\usepackage[inline]{enumitem} 
\newcommand{\corr}{(\Letter)}
\usepackage{mwe}

\usepackage{hyperref}
\usepackage{float}

\makeatletter
\renewcommand\paragraph{\@startsection{paragraph}{4}{\z@}%
{1ex \@plus 0.2ex \@minus 0.2ex}%
{-1em}%
{\normalfont\normalsize\bfseries}}
\makeatother

\begin{document}

\title{Momentum-Guided Federated Split Distillation for Personalized Temporal Edge Intelligence}

\titlerunning{Momentum-Guided Federated Split Distillation}

\author{Ahmed-Rafik Baahmed\inst{1}\orcidID{0009-0006-4313-2400} \corr \and
Jean-François Dollinger\inst{1}\orcidID{0000-0002-6688-2320} \and
Mohamed El Amine Brahmia\inst{1}\orcidID{0000-0003-0114-210X} \and
Mourad Zghal\inst{1}\orcidID{0000-0003-1329-8574}}


\authorrunning{A.-R. Baahmed et al.}

\institute{CESI, CESI LINEACT UR 7527, Strasbourg, France \email{\{abaahmed,jfdollinger,abrahmia,mzghal\}@cesi.fr}}

\maketitle              

\begin{abstract}
We propose a momentum-guided federated split distillation framework for personalized, efficient, and autonomous temporal edge intelligence.
We introduce TeRR-SAtt, our novel temporal reservoir student attention design that combines fixed reservoir representations, a lightweight temporal student, and personalized output modules. We also present AMGF, our anticipatory momentum-guided fusion mechanism that clusters clients through learning momentum and derives specialized teacher updates. On real-world smart-building data, TeRR-SAtt reduces edge training latency by $65.50\%$, inference latency by $44.70\%$, training memory usage by $18.40\%$, and inference CPU usage by $33.10\%$ over the considered baselines. At the same time, AMGF improves local learning by up to $35.31\%$ in RMSE compared to global updates.








\end{abstract}

\section{Introduction}
\label{section:introduction}
Real-world IoT ecosystems increasingly require distributed devices to perform adaptive and time-sensitive intelligence across applications such as smart buildings \cite{Mshragi2025-cz}, smart cities \cite{SmartCity-Survey,SmartCity-Survey-1}, connected health \cite{AIoTHealthcare,AIoTHealthcare-1}, industrial automation \cite{IndustrialAIoT,IndustrialAIoT-1}, and cybersecurity \cite{CyberAIIoT}.
These systems rely on heterogeneous device networks operating under complex, unstable environments and generating high-dimensional temporal data streams. 
This motivates edge AI, where learning and inference are moved closer to the data source to reduce latency and support real-time adaptation \cite{EdgeAI}. 
However, the limited computation, memory, energy, and communication resources of IoT devices make it challenging to train and deploy large temporal models directly on the edge \cite{FL-IoT-Survey}.

A broad body of recent research has examined the general concept of edge intelligence, on-device learning, efficient AI, and Federated Learning (FL) as means of deploying AI in resource-constrained IoT settings \cite{HiFEL-OCKT}.
FL is especially attractive as it enables collaborative learning without sharing raw data \cite{FL-Edge-SLR,Hyperparam-Impact}, but it typically assumes that clients can train and synchronize complete models. 
Split Learning (SL) reduces this burden by partitioning the model between weak clients and a stronger server \cite{SL-1,SL-2,Dist-ML-Survey}. Federated split learning (FSL) combines model partitioning with collaborative training \cite{FSL-1,FSL-2}. 
Nevertheless, split-based methods still face a training-deployment mismatch: the server-side component may remain necessary for inference, limiting autonomous edge deployment under latency, bandwidth, or connectivity constraints.

This limitation is amplified by client heterogeneity and temporal complexity. 
IoT clients may differ in sensing modalities, hardware capabilities, local objectives, noise patterns, and temporal distributions \cite{HiFEL-OCKT}. 
In such settings, a single global model may undermine personalization and convergence by failing to capture client-specific dynamics \cite{Heter-FL}. 
This is just as relevant in federated split learning, where a shared server-side component may still be forced to represent heterogeneous and partially incompatible local dynamics.
Thus, an important challenge remains in balancing collaborative learning and personalization, since practical IoT systems require both knowledge cooperation and local specialization.
Moreover, many edge scenarios depend on temporal context and evolving dynamics \cite{Edge-AI-Survey-Temporal}.
As such, deep temporal modeling is highly valuable in IoT ecosystems \cite{Temporal-Intelligence-IoT-1,Temporal-Intelligence-IoT-2,Temporal-Intelligence-IoT-3} to handle long-range dependencies, temporal salience, and sequence behavior more effectively compared to shallow static models.
However, temporal reasoning remains expensive for the edge \cite{Temp-Att-Expensive-1,Temp-Att-Expensive-2}.
Beyond model efficiency, heterogeneous temporal dynamics can also make instantaneous gradients noisy and unstable. This motivates update mechanisms that account for evolving client learning trajectories.

To address these challenges, we propose a momentum-guided federated split distillation framework for personalized temporal edge intelligence. At the edge side, we introduce \textbf{TeRR-SAtt}, our \textbf{Temporal ReseRvoir Student Attention} model design that combines reservoir-based temporal representation, lightweight student learning, temporal attention learning, and specialized output modules to enable personalized learning and efficient autonomous inference. The server hosts a higher-capacity temporal teacher, used only during training to provide personalized knowledge guidance without imposing server dependency at deployment. Unlike existing federated split distillation schemes that rely on a single shared server update, our approach identifies groups of clients with compatible learning trajectories and derives cluster-aware teacher updates. At the core of this process, \textbf{Anticipatory Momentum-Guided Fusion (AMGF)} provides teacher specialization and momentum-based anticipation\footnote{We refer to leveraging recent client learning trajectories to guide future learning.} to support personalized temporal knowledge transfer.

In summary, our contributions are threefold:
(i) TeRR-SAtt, our temporal reservoir student attention design that addresses the training--deployment mismatch of temporal edge learning;
(ii) our learning momentum clustering for discovering clients with compatible learning trajectories; and
(iii) our anticipatory momentum-guided fusion mechanism that jointly controls specialization and anticipatory momentum reinforcement.

\section{Related Work}

\paragraph{Federated Split Learning for Edge Intelligence.}
Recent FSL studies mainly aim to reduce computation, storage, and communication costs in edge environments.
EPSL~\cite{EPSL} improves training latency through parallel split execution over resource-constrained wireless edge networks, while CSE-FSL~\cite{FSL-2} reduces communication and storage overhead by limiting frequent gradient transmission. EUSFL~\cite{EUSFL} adopts an edge-assisted U-shaped split federated architecture with heterogeneous models to improve deployment flexibility, whereas MP-FSL~\cite{MP-FSL} applies pruning to lower resource consumption on constrained devices.
These works improve the feasibility of FSL under resource constraints, but treat the server-side model as globally shared rather than adaptively personalized to clients' learning requirements.

\paragraph{Federated Distillation and Personalization.}
Knowledge distillation has been widely used to improve communication efficiency, support model heterogeneity, and enable personalization in federated learning. 
FSMKD \cite{FSMKD} combines federated split learning with mutual knowledge distillation through a two-body structure that supports personalized local models and a shared server-side model. 
PFL-DKD \cite{PFL-DKD} and DKD-pFed \cite{DKD-pFed} use decoupled knowledge distillation to improve personalized FL under heterogeneous distributions, while pFedVS \cite{pFedVS} adopts self-distillation for personalized vehicular edge learning to improve convergence and reduce the negative effect of client selection. 
These studies confirm the importance of distillation and personalization, but do not address how a server-side temporal teacher should adapt to compatible or conflicting client learning trajectories.
Unlike our prior HiFEL-OCKT \cite{HiFEL-OCKT}, where edge devices share a common trainable backbone, the present framework keeps all deployed modules client-specific and delegates temporal reasoning to a training-only server teacher via distillation and momentum-guided collaboration.

\paragraph{Client Clustering and Gradient-Based Aggregation.}
A growing line of work addresses non-I.I.D. federated learning through clustering, adaptive aggregation, or momentum-based optimization.
Cheng et al. \cite{ICLR2024_291d92e9} showed that momentum can provably benefit non-I.I.D. FL by improving FedAvg and SCAFFOLD without relying on bounded heterogeneity assumptions, while Li et al. proposed FedJSCM \cite{FedJSCM}, a joint server-client momentum method to mitigate data heterogeneity.
However, in these methods, momentum remains primarily an optimization correction mechanism for a single global model and does not address personalization, collaboration-group discovery, or cluster-specialized collaborative adaptation.
Other methods, such as StoCFL \cite{StoCFL} and PFedCSCBO \cite{PFedCSCBO}, form client clusters using distributional, stochastic, or classifier-similarity criteria, but they do not exploit momentum for future collaboration discovery.
In contrast, AMGF uses learning-path momentum to discover compatible clients and construct personalized updates.

Overall, existing FSL and distillation methods mainly address computation placement and knowledge transfer, while clustering and aggregation methods focus on update combination. AMGF bridges these directions by determining which clients should share a teacher update and how this update should be specialized and anticipated based on their evolving learning trajectories.

\section{Methodology}
This section presents our proposed momentum-guided federated split distillation framework.
We first formulate the personalized temporal edge learning problem, then describe our TeRR-SAtt model design, and finally introduce AMGF to discover trajectory-compatible client groups and construct personalized updates.

\subsection{Problem Statement}
We consider a distributed IoT ecosystem with $N$ resource-constrained clients:
\begin{equation}
    \mathcal{N} = \{1,2,\dots,N\},
    \label{eq:clients-set}
\end{equation}
where each client $i \in \mathcal{N}$ owns a private temporal dataset:
\begin{equation}
    \mathcal{D}_{i} = \{(\mathbf{X}_{i}^{j}, \mathbf{y}_{i}^{j})\}_{j=1}^{n_i},
    \label{eq:client-dataset}
\end{equation}
with $\mathbf{X}_i^j \in \mathbb{R}^{s_i \times f_i}$ denoting a multivariate time-series sample of length $s_i$ with $f_i$ features, and $\mathbf{y}_i^j$ denoting the corresponding prediction target.
Clients may differ in sensing modalities, temporal distributions, local objectives, noise patterns, and resource budgets, leading to non-I.I.D. temporal learning conditions.

The objective is to learn a personalized deployable temporal predictor $\mathcal{F}_{i}$ for each client, while preserving data locality and satisfying client-side resource constraints.
Formally, the personalized edge learning objective is written as:
\begin{equation}
    \min_{\{\mathcal{F}_{i}\}_{i=1}^{N}}
    \sum_{i=1}^{N}
    \mathbb{E}_{(\mathbf{X}_i,\mathbf{y}_i)\sim \mathcal{D}_i}
    \Bigl[
    \ell
    \bigl(
    \mathcal{F}_{i}(\mathbf{X}_i), \mathbf{y}_i
    \bigr)
    \Bigr],
    \quad
    \text{s.t.} \quad \mathrm{Cost}(\mathcal{F}_{i}) \leq B_i, \forall i \in \mathcal{N},
    \label{eq:local-client-objective}
\end{equation}
where $\ell(\cdot)$ denotes the prediction loss, $\mathrm{Cost}(\mathcal{F}_{i})$ is the client-side computational and memory cost of deploying $\mathcal{F}_{i}$, and $B_i$ is the resource budget of client $i$.
Solving Eq.~\eqref{eq:local-client-objective} purely locally enables autonomy but ignores useful cross-client temporal knowledge. Conversely, enforcing a single global model or a single globally shared server-side update may be inadequate under heterogeneous temporal dynamics, since clients can follow incompatible learning directions. This issue is particularly important in split learning settings, where the server-side temporal component may receive learning signals induced by clients with divergent learning objectives.

To exploit collaboration without compromising autonomous deployment, we consider a training-only temporal guidance mechanism. During training round $t$, each client induces a learning signal, denoted by the client gradients vector $\mathbf{g}^{t}_{i}$, which reflects how client $i$ would adapt the shared model according to its local data. A naive global update would aggregate these signals as:
\begin{equation}
    \mathbf{g}^t = \sum_{i=1}^{N} w_i \mathbf{g}_i^t,
    \label{eq:global-teacher-update}
\end{equation}
where $w_i$ denotes the aggregation weight of client $i$. However, such averaging can suppress useful group-specific temporal knowledge when clients exhibit conflicting or only partially compatible learning trajectories. This may reduce personalization, slow convergence, and require additional local refinement on constrained edge devices.
Therefore, the central methodological problem addressed in this work is the following:
Given client-induced learning signals $\{\mathbf{g}_i^t\}_{i=1}^{N}$ across rounds, how can the server discover clients with compatible learning trajectories and derive personalized guidance updates that balance specialization, consistency, and anticipation while preserving efficiency and autonomous inference?

This formulation separates two complementary requirements: a lightweight client model that remains server-independent at inference time but efficient and effective at training time, and adaptive server-side guidance that follows heterogeneous client learning trajectories rather than a single global update. TeRR-SAtt addresses the first requirement, while AMGF addresses the second.

\subsection{TeRR-SAtt Model Design}
To instantiate the deployable predictor $\mathcal{F}_{i}$ in Eq.~\eqref{eq:local-client-objective}, we propose our \textbf{Temporal ReseRvoir Student Attention (TeRR-SAtt)} design.
TeRR-SAtt follows a U-shaped split-distillation structure in which the client retains both the input-side representation module and the output-side prediction module, while the server hosts a higher-capacity temporal teacher that is used only during training.

For each client $i$, the deployable edge model is defined as:
\begin{equation}
    \mathcal{F}_{i}(\mathbf{X}_i)
    =
    \mathcal{O}_i
    \left(
    \mathcal{S}_i
    \left(
    \mathcal{E}_i(\mathbf{X}_i)
    \right)
    \right),
    \label{eq}
\end{equation}
where $\mathcal{E}_i$ is a lightweight reservoir-based temporal representation module, $\mathcal{S}_i$ is a compact temporal student module, and $\mathcal{O}_i$ is a personalized output module. This decomposition allows the client to retain an autonomous inference path after training, while keeping the trainable temporal component lightweight enough for resource-constrained edge deployment.

\begin{figure}[htbp]
\vspace{-5mm}
    \centering
    \includegraphics[width=\linewidth]{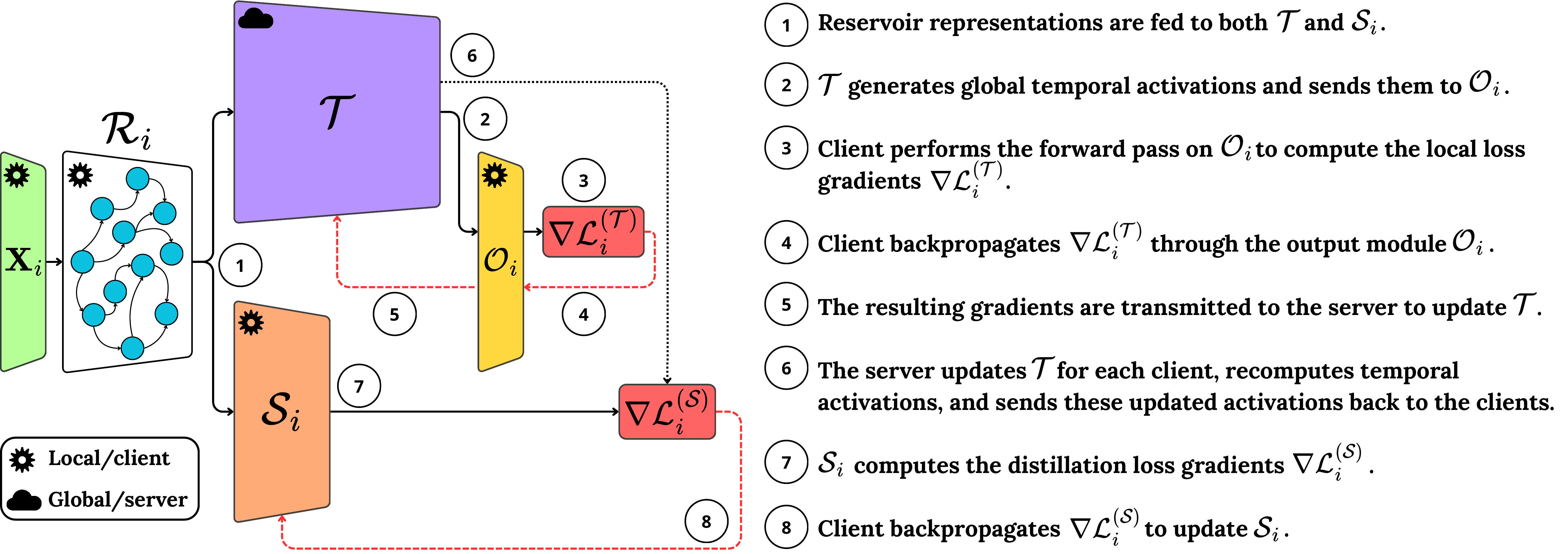}
    \caption{TeRR-SAtt design.}
    \label{fig:purpl_kfd_model_design}
\vspace{-5mm}
\end{figure}

As shown in Fig.~\ref{fig:purpl_kfd_model_design}, the reservoir representation module first maps the input sequence into an expanded temporal representation:
\begin{equation}
    \mathcal{R}_i = \mathcal{E}_i(\mathbf{X}_i),
    \label{eq:reservoir-representation}
\end{equation}
where $\mathcal{R}_i$ denotes the client-side reservoir state representation. Since the reservoir dynamics are fixed, $\mathcal{E}_i$ provides temporal expansion at low training cost. This makes it suitable for weak IoT clients that cannot efficiently train large recurrent or attention-based temporal encoders from scratch.
During training, the same representation $\mathcal{R}_i$ is processed by a server with a high-capacity temporal teacher $\mathcal{T}$.
The teacher produces the temporal activation $\mathbf{H}^{(\mathcal{T})}_i$ which is transmitted back to client $i$ and passed through the local output module:
\begin{equation}
    \widehat{\mathbf{y}}^{(\mathcal{T})}_i
    =
    \mathcal{O}_i
    \left(
    \mathbf{H}^{(\mathcal{T})}_i
    \right),
    \qquad
    \text{s.t.} \quad \mathbf{H}^{(\mathcal{T})}_i = \mathcal{T}(\mathcal{R}_i).
    \label{eq:output-predictions}
\end{equation}
The corresponding supervised loss is:
\begin{equation}
    \mathcal{L}^{(\mathcal{T})}_i
    =
    \ell
    \left(
    \widehat{\mathbf{y}}^{(\mathcal{T})}_i,
    \mathbf{y}_i
    \right).
    \label{eq:teacher-assisted-pred-loss}
\end{equation}

The client backpropagates the gradient of this loss through its local output module and obtains the gradient with respect to the teacher activation:
\begin{equation}
    \mathbf{u}^{t}_{i}
    =
    \frac{\partial \mathcal{L}^{(\mathcal{T})}_i}
    {\partial \mathbf{H}^{(\mathcal{T})}_i}.
    \label{eq:gradient-with-respect-to-teacher}
\end{equation}
The gradient $\mathbf{u}^{t}_{i}$ is transmitted to the server, where it induces a teacher-side gradient:
\begin{equation}
    \mathbf{g}^{t}_{i}
    =
    - \nabla{\Theta_\mathcal{T}}
    \mathcal{L}^{(\mathcal{T})}_i,
    \label{eq:teacher-gradients}
\end{equation}
with $\Theta_\mathcal{T}$ denoting the parameters of the server-side teacher. Instead of updating the teacher using naive aggregation, these client-induced teacher gradients are later processed by our AMGF mechanism to derive specialized teacher updates.

After the teacher update, the server recomputes the temporal activation for client $i$ using the updated teacher variant assigned to its collaboration group:
\begin{equation}
    \widetilde{\mathbf{H}}^{(\mathcal{T})}_i
    =
    \mathcal{T}^{+}_{k(i,t)}
    \left(
    \mathcal{R}_i
    \right),
    \label{eq:updated-teacher-temp-activ}
\end{equation}
where $\mathcal{T}^{+}_{k(i,t)}$ denotes the updated teacher associated with the collaboration group of client $i$ at round $t$.
This updated activation is returned to the client and used as a distillation target for the local temporal student. The student produces:
\begin{equation}
    \mathbf{H}^{(\mathcal{S})}_i
    =
    \mathcal{S}_i
    \left(
    \mathcal{R}_i
    \right),
    \label{eq:student-temp-activations}
\end{equation}
and is trained by minimizing the representation-level distillation loss:
\begin{equation}
    \mathcal{L}^{(\mathcal{S})}_i
    =
    d
    \left(
    \mathbf{H}^{(\mathcal{S})}_i,
    \widetilde{\mathbf{H}}^{(\mathcal{T})}_i
    \right),
    \label{eq:student-temp-loss}
\end{equation}
where $d(\cdot,\cdot)$ denotes a representation-matching loss. The parameters of $\mathcal{S}_i$ are then updated locally by backpropagating $\mathcal{L}^{(\mathcal{S})}_i$.

Through this design, TeRR-SAtt separates training-time collaborative temporal reasoning from inference-time edge autonomy. During training, the server teacher provides high-capacity temporal guidance that improves the local student. During inference, each client predicts using only:
\begin{equation}
    \widehat{\mathbf{y}}_i
    =
    \mathcal{O}_i
    \left(
    \mathcal{S}_i
    \left(
    \mathcal{E}_i(\mathbf{X}_i)
    \right)
    \right).
    \label{eq:inference-formula}
\end{equation}
Thus, TeRR-SAtt addresses the training--deployment mismatch of federated split learning by combining training-time collaborative guidance with efficient autonomous edge inference.
Since it does not specify how the server-side teacher should adapt across heterogeneous clients, we next introduce AMGF, which uses client learning trajectories to discover compatible collaboration groups and construct personalized teacher updates.

\subsection{Anticipatory Momentum-guided Fusion}
We introduce \textbf{Anticipatory Momentum-Guided Fusion (AMGF)} to adapt the server-side teacher across heterogeneous clients without relying on a single global update. AMGF uses the client-induced teacher gradients defined in Eq.~\eqref{eq:teacher-gradients} to identify clients with compatible learning trajectories and to construct cluster-specialized teacher variants. Let $\boldsymbol{\theta}_{\mathcal{T}^{0}}$ denote the vectorized parameters of the initial teacher. At round $t$, each client $i$ receives teacher guidance from a cluster-specific teacher variant and induces a teacher-side gradient $\mathbf{g}_{i}^{t}$. 
Since clients may receive teacher guidance from different cluster-specific variants across rounds, directly comparing instantaneous gradients can be misleading. To express all client requests in a common coordinate system, AMGF maintains, for each client, a teacher-relative path:
\begin{equation}
    \mathbf{p}_{i}^{t}
    =
    \boldsymbol{\theta}_{\mathcal{T},i}^{t}
    -
    \boldsymbol{\theta}_{\mathcal{T}}^{0},
    \label{eq:path-aware-teacher-path}
\end{equation}
where $\boldsymbol{\theta}_{\mathcal{T},i}^{t}$ is the teacher variant used by client $i$ at the beginning of round $t$. Thus, $\mathbf{p}_{i}^{t}$ represents the displacement already experienced by client $i$ with respect to the initial teacher. The effective path-aware update request is defined as:
\begin{equation}
    \mathbf{q}_{i}^{t}
    =
    \mathbf{p}_{i}^{t}
    + \mathbf{g}_{i}^{t}.
    \label{eq:path-aware-effective-gradient}
\end{equation}

To reduce the effect of noisy instantaneous gradients under non-I.I.D. temporal data, AMGF maintains a smoothed path-aware momentum:
\begin{equation}
    \mathbf{m}_{i}^{t}
    =
    \begin{cases}
        \mathbf{q}_{i}^{1}, & \text{if } t=1, \\[4pt]
        \beta \mathbf{m}_{i}^{t-1}
        +
        (1-\beta)\mathbf{q}_{i}^{t}, & \text{if } t>1,
    \end{cases}
    \label{eq:momentum-equation}
\end{equation}
where $\beta \in [0,1]$ controls temporal smoothing. Unlike standard momentum applied only to raw gradients, $\mathbf{m}_{i}^{t}$ summarizes the persistent teacher-relative displacement tendency requested by client $i$.

To directly capture compatibility between client update requests, we construct a pairwise momentum-affinity matrix:
\begin{equation}
    \mathcal{A}_{ij}^{t}
    =
    \operatorname{cos}
    \left(
    \mathbf{m}_{i}^{t},
    \mathbf{m}_{j}^{t}
    \right),
    \label{eq:pairwise-momentum-similarity}
\end{equation}
where $\mathcal{A}_{ij}^{t} \in [-1,1]$ measures the directional compatibility between the path-aware momentum vectors of clients $i$ and $j$. Positive values indicate aligned learning trajectories, near-zero values indicate weak relations, and negative values indicate conflicting update directions.
Using $\mathcal{A}^{t}$ as the similarity matrix, AMGF applies \textit{Affinity Propagation Clustering}~\cite{Aff-Prop-Clustering} to obtain $\mathcal{G}^{t} = \{G_{1}^{t},G_{2}^{t},\dots,G_{K^{t}}^{t}\}$,
where $K^{t}$ is automatically inferred at round $t$. This allows the number of collaboration groups to adapt dynamically as client learning trajectories evolve.
Each group $G_k^t$ gathers clients with strongly aligned momentum directions, reflecting compatible teacher-relative learning trajectories that can benefit from a shared teacher specialization.
Thus, for each cluster $G_{k}^{t}$, AMGF computes a cluster-level momentum:
\begin{equation}
    \mathbf{m}_{[k]}^{t}
    =
    \sum_{i \in G_{k}^{t}}
    \bar{w}_{i,k}^{t}
    \mathbf{m}_{i}^{t},
    \qquad
    \bar{w}_{i,k}^{t}
    =
    \frac{w_i}
    {\sum_{j \in G_{k}^{t}} w_j},
    \label{eq:cluster-gradient-momentum}
\end{equation}
where $\bar{w}_{i,k}^{t}$ is the normalized contribution of client $i$ within cluster $G_{k}^{t}$, and $\mathbf{m}_{[k]}^{t}$ captures the smoothed and unified displacement request for the entire cluster.
Path-aware trajectory intuition is illustrated in App.~\ref{app:amgf-geometric-intuition}.

To further accelerate convergence and prevent optimization oscillations, we leverage these smoothed trajectories to inject \textit{learning anticipation} into the cluster-specific teacher update.
We introduce an adaptive anticipation factor to avoid using a fixed look-ahead step. For each cluster, we measure the intra-cluster directional agreement:
\begin{equation}
    A_{[k]}^{t}
    =
    \begin{cases}
    1, & \text{if } |G_{k}^{t}|=1, \\[4pt]
    \dfrac{2}{|G_{k}^{t}|(|G_{k}^{t}|-1)}
    \displaystyle\sum_{\substack{i<j\ i,j \in G_{k}^{t}}}
    \max
    \left(
    0,
    \operatorname{cos}
    \left(
    \mathbf{m}_{i}^{t},
    \mathbf{m}_{j}^{t}
    \right)
    \right),
    & \text{otherwise},
    \end{cases}
    \label{eq:cluster-agreement}
\end{equation}
and the temporal consistency of client trajectories:
\begin{equation}
    C_{[k]}^{t}
    =
    \begin{cases}
    0, & \text{if } t=1, \\[4pt]
    \dfrac{1}{|G_{k}^{t}|}
    \displaystyle\sum_{i \in G_{k}^{t}}
    \max
    \left(
    0,
    \operatorname{cos}
    \left(
    \mathbf{m}_{i}^{t},
    \mathbf{m}_{i}^{t-1}
    \right)
    \right),
    & \text{if } t>1.
    \end{cases}
    \label{eq:trajectory-stability}
\end{equation}
Both $A_{[k]}^{t}$ and $C_{[k]}^{t}$ are naturally bounded within $[0,1]$. The adaptive anticipation coefficient is therefore defined as:
\begin{equation}
    \alpha_{[k]}^{t}
    =
    \alpha_{\max}
    A_{[k]}^{t}
    C_{[k]}^{t},
    \label{eq:alpha-equation}
\end{equation}
where $\alpha_{\max}$ sets the maximum look-ahead magnitude.
This multiplicative gate does not impose a hard activation threshold. Instead, the reliability of anticipation is encoded continuously by the product $A^{t}_{[k]}C^{t}_{[k]}$ in Eq.~\eqref{eq:alpha-equation}. When either intra-cluster agreement or temporal consistency is low, $\alpha^{t}_{[k]}$ becomes small and anticipation is naturally suppressed; when both are high, $\alpha^{t}_{[k]}$ approaches $\alpha_{\max}$. Thus, anticipation is not a binary decision, but a reliability-gated continuous look-ahead term.
A behavioral analysis is provided in App.~\ref{app:amgf-behavior-appendix}.

The cluster-specialized teacher is finally constructed as:
\begin{equation}
    \boldsymbol{\theta}_{\mathcal{T},[k]}^{t+1}
    =
    \boldsymbol{\theta}_{\mathcal{T}}^{0}
    +
    \left(
    \eta + \alpha_{[k]}^{t}
    \right)
    \mathbf{m}_{[k]}^{t},
    \label{eq:cluster-teacher-update}
\end{equation}
where $\eta$ is the base teacher-update scaling factor. Expressing the teacher update relative to $\boldsymbol{\theta}_{\mathcal{T}}^{0}$ keeps all cluster-specific teachers in a shared reference coordinate system and limits uncontrolled recursive drift. The resulting teacher variant $\mathcal{T}_{[k]}^{t+1}$ is then used to recompute the temporal activations of clients in $G_{k}^{t}$, as in Eq.~\eqref{eq:updated-teacher-temp-activ}. These activations are returned to the clients and used as distillation targets for the local students through Eq.~\eqref{eq:student-temp-loss}.
Bounded drift is discussed in App.~\ref{app:stability-bounded-drift}.

At the end of the round, the path variable of each client is updated according to the teacher variant assigned to its current cluster:
\begin{equation}
    \mathbf{p}_{i}^{t+1}
    =
    \boldsymbol{\theta}_{\mathcal{T},[k(i,t)]}^{t+1}
    -
    \boldsymbol{\theta}_{\mathcal{T}}^{0},
    \quad
    \forall i \in G_{k(i,t)}^{t}.
    \label{eq:delta-extraction}
\end{equation}
In this way, AMGF provides personalized teacher guidance by grouping clients according to compatible learning trajectories and
continuously scaling anticipation according to the reliability of the cluster direction.

\section{Experimental Results \& Discussion}
The experimental evaluation is conducted on a smart-building forecasting use case using the LBNL building dataset \cite{LBNL-Dataset}. We use 20 edge devices, each corresponding to a thermal zone, with three shared sensing/control features: air fan speed (\%), temperature ($^\circ$F), and heating-water valve position (\%). Each device performs local multi-step forecasting from a $72$-step input window to the next $6$ steps. This setting reflects a realistic smart-building edge scenario where distributed zones exhibit heterogeneous temporal dynamics.

\paragraph{Edge Computational Efficiency of TeRR-SAtt.}
We first evaluate the edge-side computational efficiency of TeRR-SAtt against two reference deployments: Traditional FL and Traditional FSL.
All models process normalized multivariate data with a batch size of $16$ for $87$ model update iterations on a Raspberry Pi~$5$ with $8$~GB memory.
Traditional FL deploys the full temporal model at the edge: a two-layer GRU encoder with $64$ and $32$ units, two multi-head self-attention blocks with $4$ heads, key dimension $32$, feed-forward dimension $128$, and global-average pooling, followed by a $128$-unit dense output module. Traditional FSL keeps the same edge encoder and output module at the edge but offloads the temporal attention module to the server. TeRR-SAtt instead uses a fixed sparse reservoir of size $64$ with sparsity $0.95$, spectral radius $0.9$, and leak rate $0.3$, followed by a $64$-unit GRU student and the same output module.

\begin{figure}[htbp]
    \centering
    \begin{subfigure}[b]{0.45\textwidth}
        \centering
        \includegraphics[width=\textwidth]{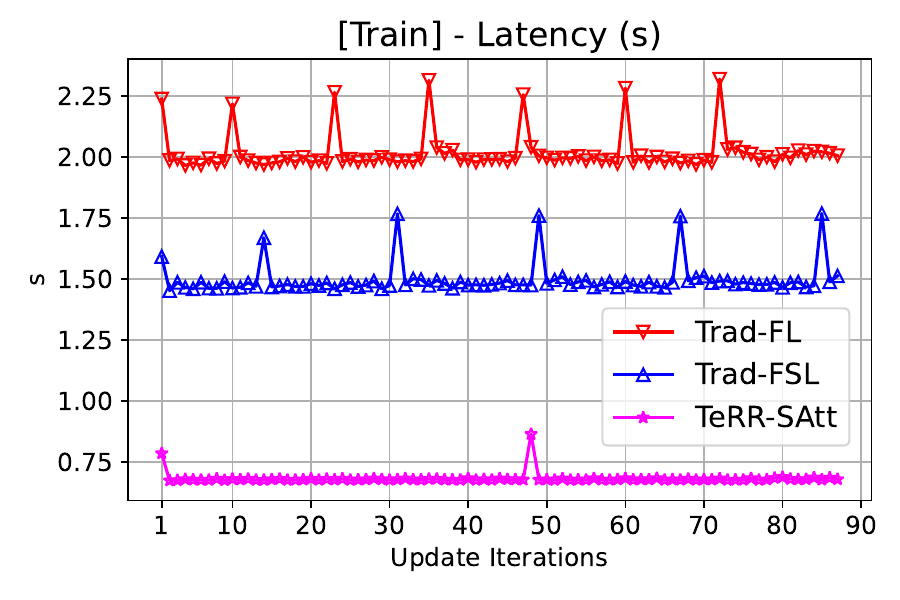}
    \end{subfigure}
    \begin{subfigure}[b]{0.45\textwidth}
        \centering
        \includegraphics[width=\textwidth]{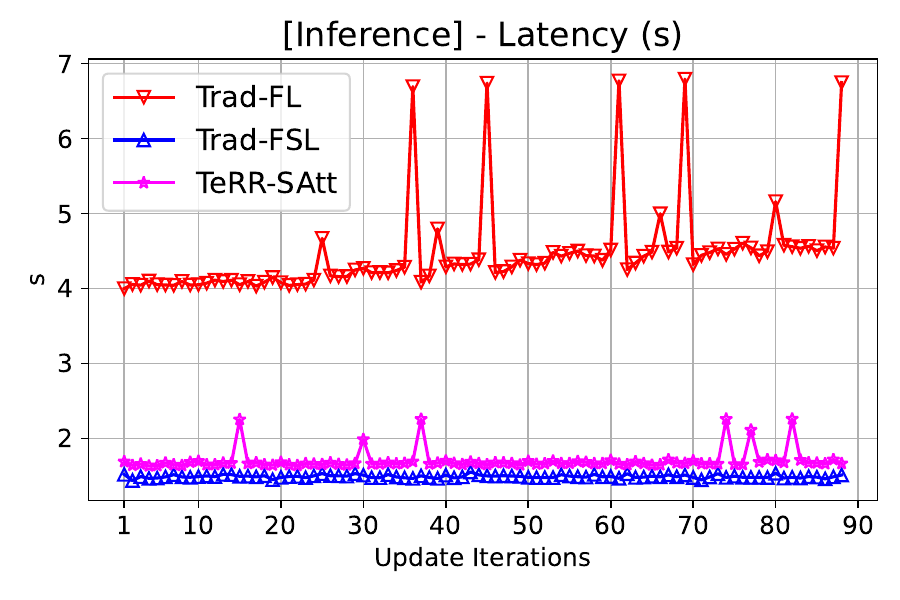}
    \end{subfigure}
    \vspace{-5mm}
    \caption{Edge-side latency comparison.}
    \label{fig:edge_train_inference_latency}
\vspace{-7mm}
\end{figure}

Fig.~\ref{fig:edge_train_inference_latency} shows that TeRR-SAtt substantially reduces training latency compared with both baselines. 
This corresponds to a latency reduction of about $65.50\%$ compared with FL and $53.70\%$ compared with FSL, i.e., TeRR-SAtt is approximately $2.9\times$ faster than FL and $2.2\times$ faster than FSL during edge-side training.
For inference, TeRR-SAtt provides the strongest deployable solution. Compared with FL, it reduces latency
by approximately $44.70\%$. More importantly, TeRR-SAtt is also superior to FSL in real end-to-end inference. The plotted FSL value reports only local edge computation, excluding the mandatory communication with the server and the server-side execution of the offloaded temporal attention module. Consequently, FSL necessarily incurs higher effective latency, whereas TeRR-SAtt performs the complete inference path locally through the reservoir, student, and output modules. Thus, TeRR-SAtt combines lower end-to-end inference latency with full edge autonomy.

\begin{figure}[htbp]
\vspace{-7mm}
    \centering
    \begin{subfigure}[b]{0.45\textwidth}
        \centering
        \includegraphics[width=\textwidth]{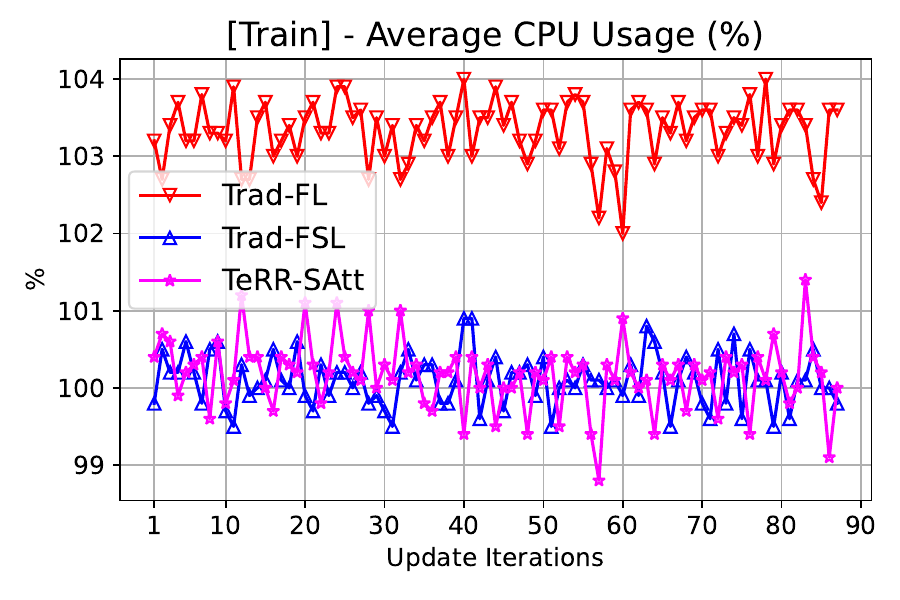}
    \end{subfigure}
    \begin{subfigure}[b]{0.45\textwidth}
        \centering
        \includegraphics[width=\textwidth]{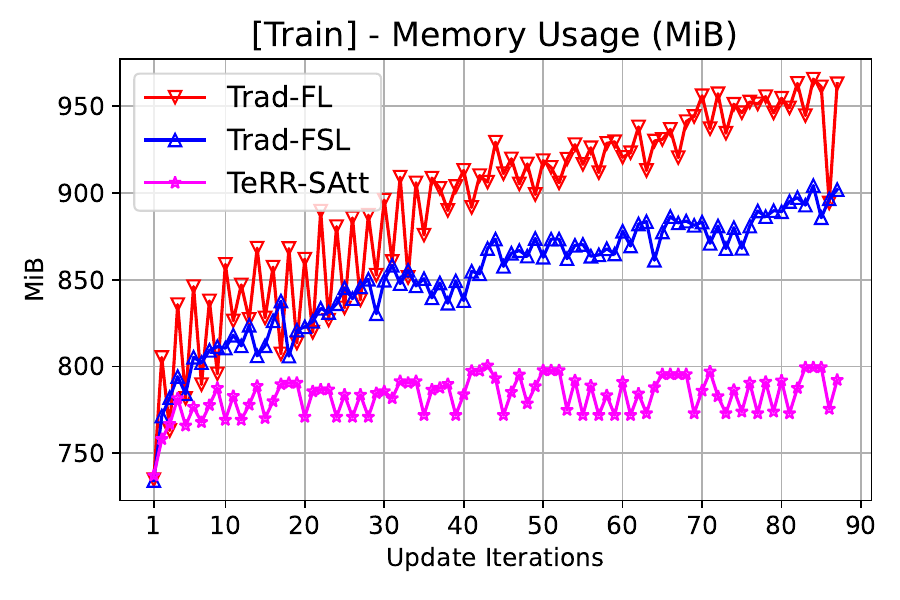}
    \end{subfigure}
    \begin{subfigure}[b]{0.45\textwidth}
        \centering
        \includegraphics[width=\textwidth]{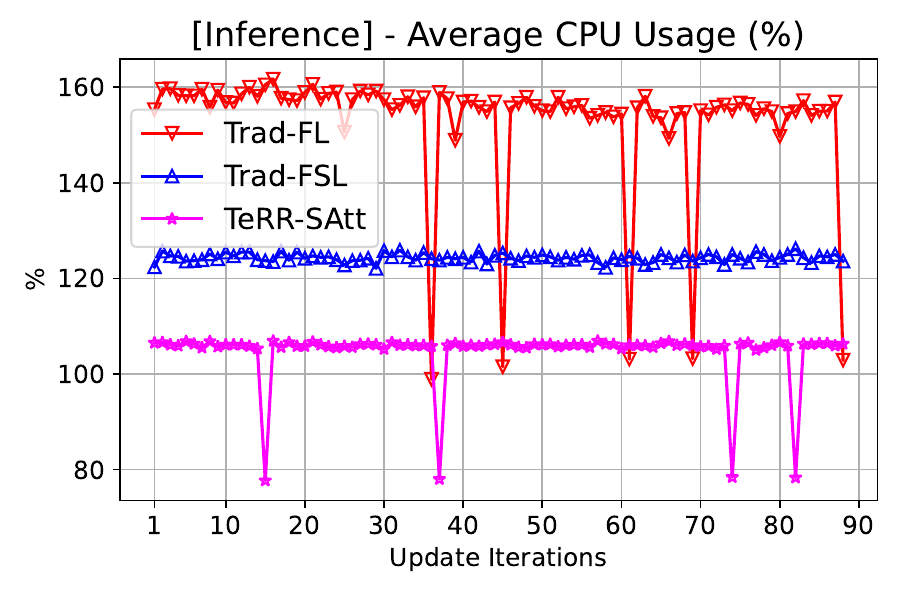}
    \end{subfigure}
    \begin{subfigure}[b]{0.45\textwidth}
        \centering
        \includegraphics[width=\textwidth]{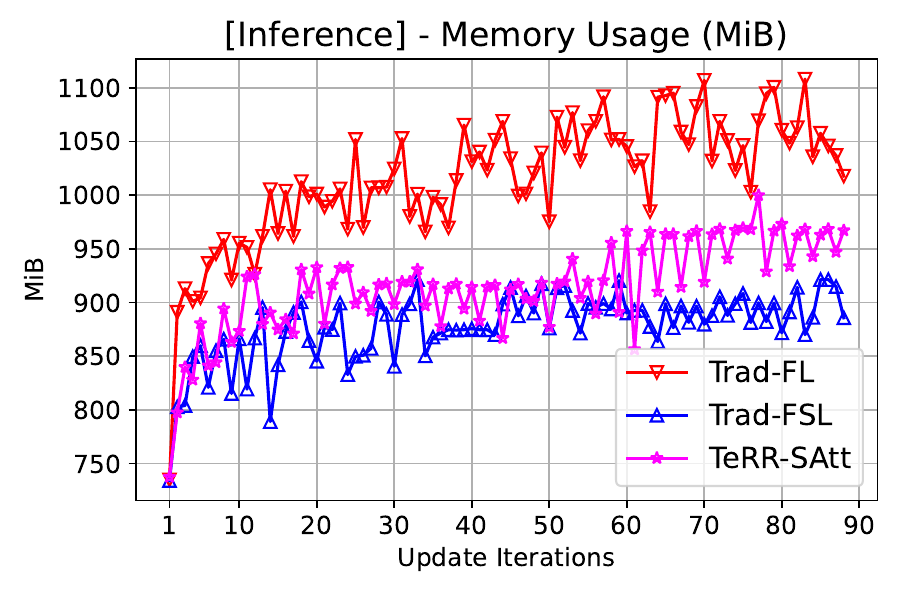}
    \end{subfigure}
    \vspace{-5mm}
    \caption{Edge-side computing-resource comparison.}
    \label{fig:edge_train_inference_performance}
\vspace{-7mm}
\end{figure}

Fig.~\ref{fig:edge_train_inference_performance} further shows that TeRR-SAtt does not increase training CPU usage compared with Traditional FSL, while Traditional FL reaches higher CPU usage. This is important because TeRR-SAtt adds a student distillation step, yet this additional learning operation does not translate into higher CPU pressure. The reason is that the reservoir module is fixed and non-trainable, while the student is significantly lighter than the full temporal attention module. As a result, TeRR-SAtt supports distillation-based learning without imposing additional edge-side computational stress. Moreover, during training, TeRR-SAtt achieves memory reductions of roughly $18.40\%$ and $11.40\%$ compared to FL and FSL, respectively. This gain is particularly relevant for constrained IoT devices, since memory usage during training can directly limit the feasibility of edge deployment.
The same trend is confirmed for inference. TeRR-SAtt achieves CPU usage reductions of approximately $33.10\%$ and $14.50\%$ compared to FL and FSL, respectively. In terms of inference memory, TeRR-SAtt remains much lighter than Traditional FL, reducing memory usage by about $10\%$.
TeRR-SAtt keeps inference memory close to the locally measured FSL baseline despite deploying a full local inference pipeline, while removing FSL's server dependency.

Overall, these results show that TeRR-SAtt achieves a stronger efficiency--autonomy trade-off than both Traditional FL and Traditional FSL.

\paragraph{AMGF Learning Effectiveness and Ablation.}
We evaluate learning performance under a unified walk-forward validation protocol with a sliding window over $10$ rounds, reporting the total RMSE evolution of representative edge nodes among the $20$ LBNL clients.
We compare four configurations:
\begin{enumerate*}[label=(\roman*)]
    \item Trad-FL/FSL\footnote{Trad-FSL partitions the same full-capacity model across edge and server; thus, its synchronous split training with full-model aggregation yields learning dynamics identical to Trad-FL; we report a single curve labeled Trad-FL/FSL.}, serving as the full-capacity globally shared reference;
    \item FSL-KD, the closest distillation baseline under the TeRR-SAtt client architecture, which uses standard global teacher updates defined by Eq.~\eqref{eq:global-teacher-update};
    \item AMGF without anticipation ($\alpha_{\max} = 0$); and
    \item full AMGF with $\beta=0.4$, $\eta=0.1$, and $\alpha_{\max} = 0.2$.
\end{enumerate*}
To complement the efficiency findings, with TeRR-SAtt's computational gains previously established (Figs.~\ref{fig:edge_train_inference_latency}--\ref{fig:edge_train_inference_performance}), we now assess its learning effectiveness relative to the full-capacity model via Trad-FL/FSL. The comparison between FSL-KD and the two AMGF variants, in turn, isolates the contribution of momentum-guided teacher specialization and adaptive anticipation under TeRR-SAtt.

\begin{figure}[htbp]
\vspace{-8mm}
    \centering
    \begin{subfigure}[b]{0.325\textwidth}
        \centering
        \includegraphics[width=1.05\textwidth]{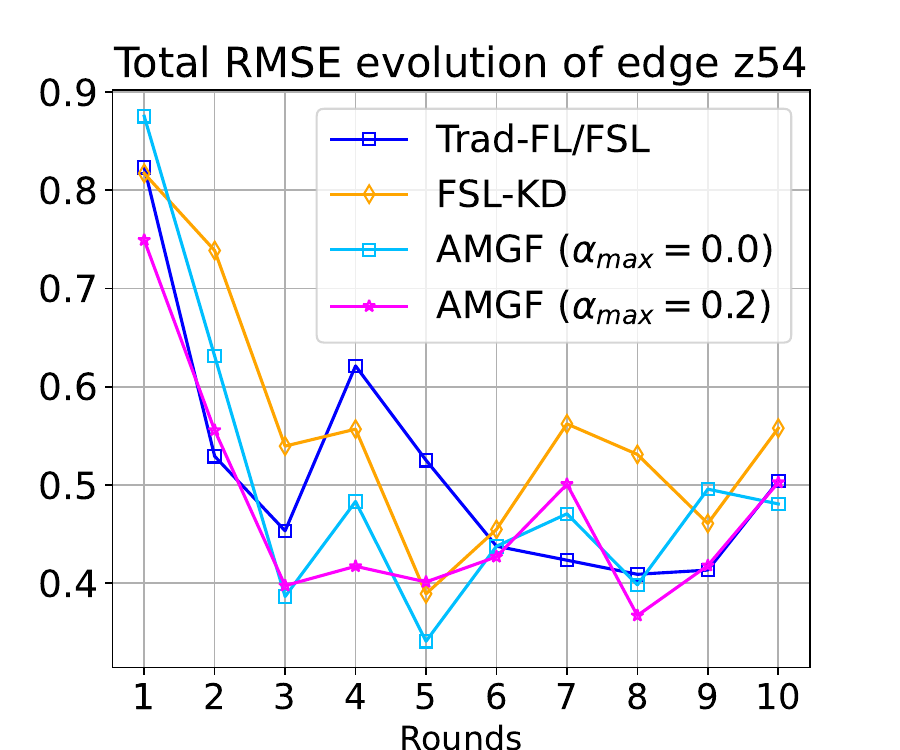}
    \end{subfigure}
    \begin{subfigure}[b]{0.325\textwidth}
        \centering
        \includegraphics[width=1.05\textwidth]{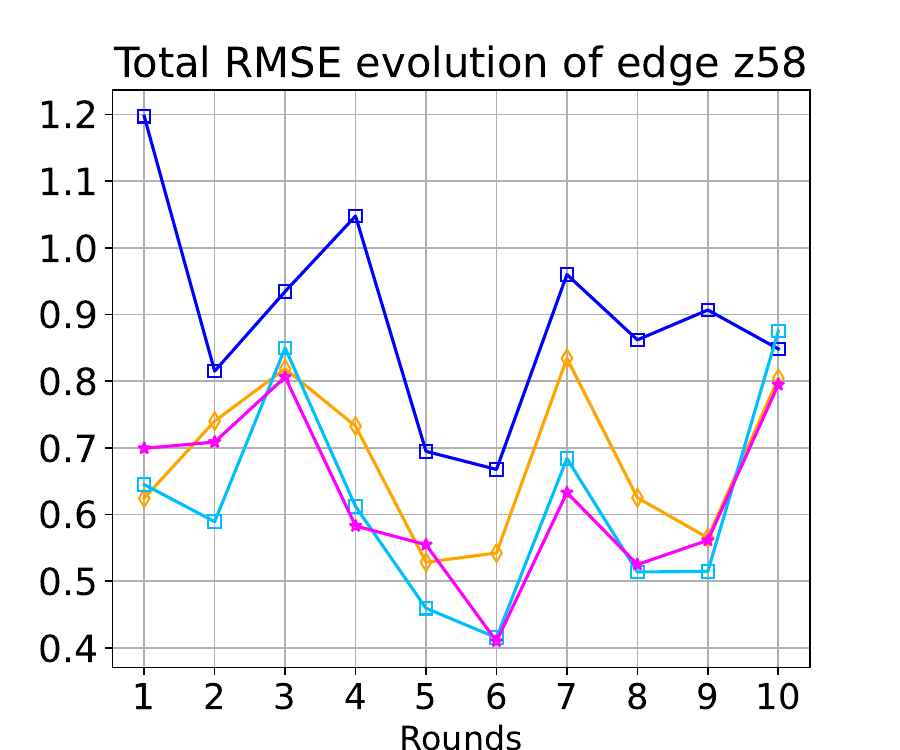}
    \end{subfigure}
    \begin{subfigure}[b]{0.325\textwidth}
        \centering
        \includegraphics[width=1.05\textwidth]{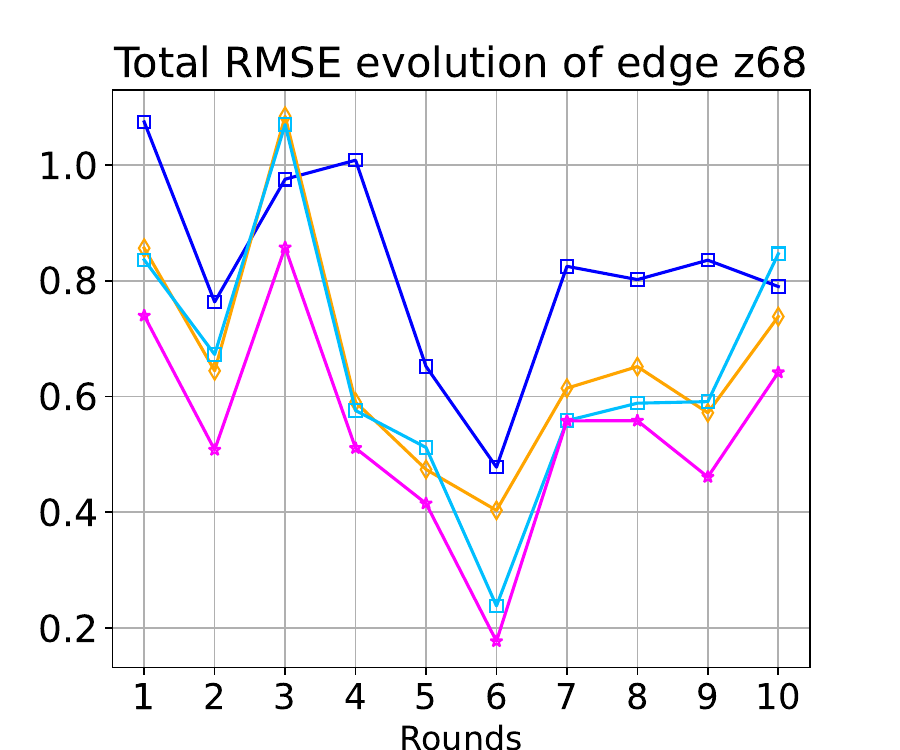}
    \end{subfigure}
    \begin{subfigure}[b]{0.325\textwidth}
        \centering
        \includegraphics[width=1.05\textwidth]{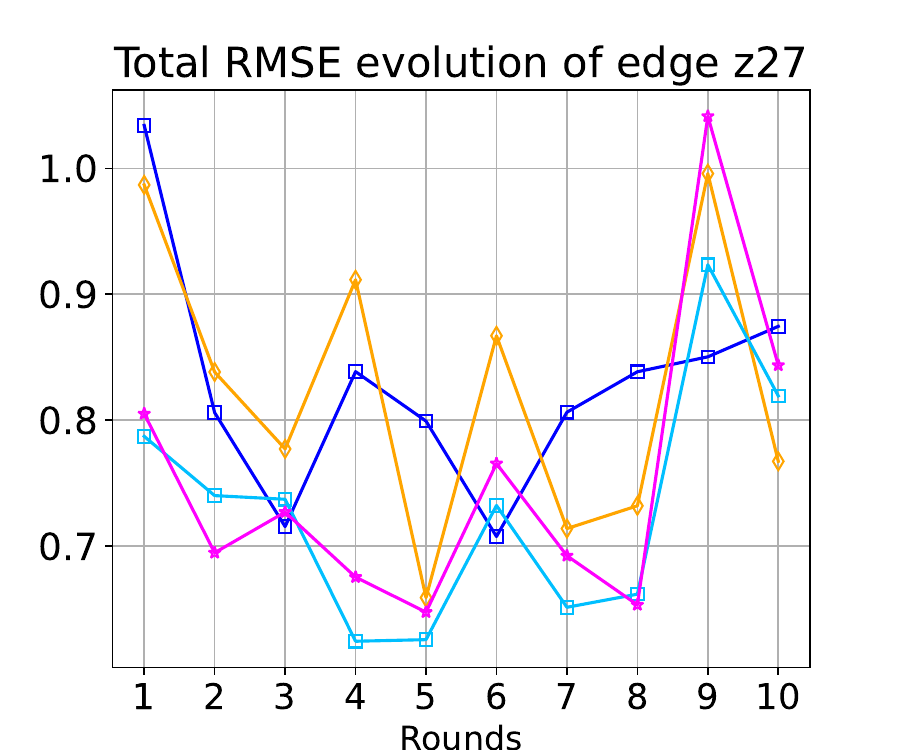}
    \end{subfigure}
    \begin{subfigure}[b]{0.325\textwidth}
        \centering
        \includegraphics[width=1.05\textwidth]{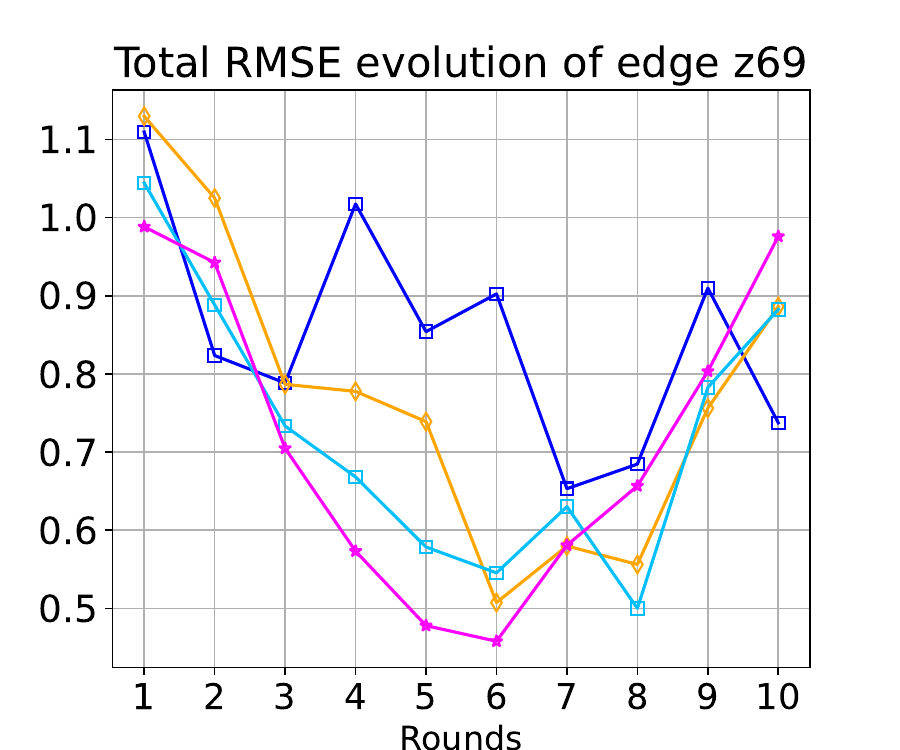}
    \end{subfigure}
    \begin{subfigure}[b]{0.325\textwidth}
        \centering
        \includegraphics[width=1.05\textwidth]{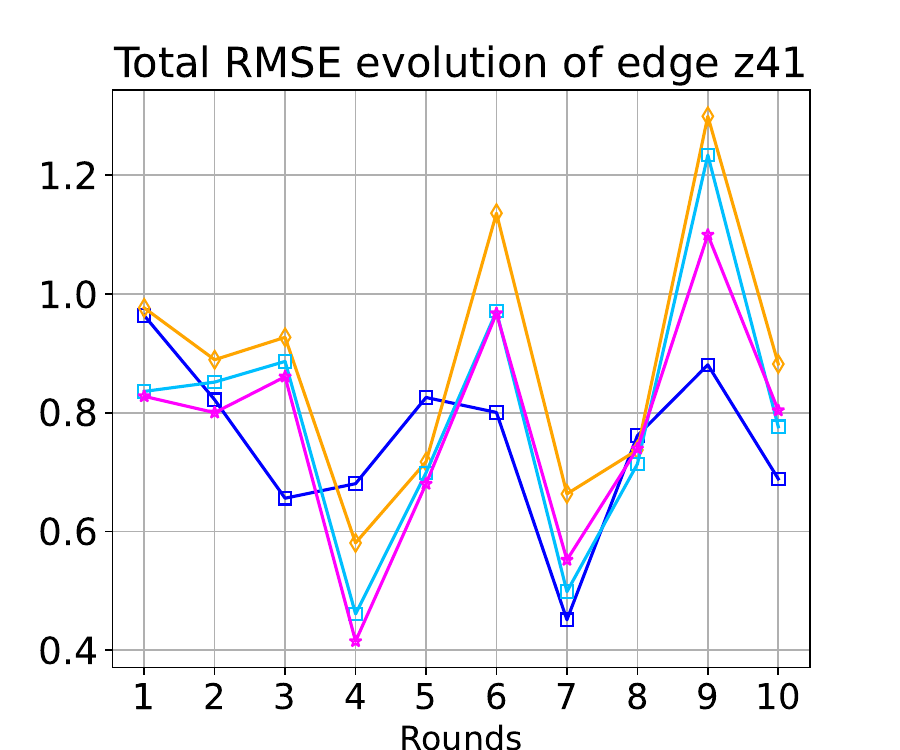}
    \end{subfigure}
    \vspace{-2mm}
    \caption{Learning performance comparison across representative edges.}
    \label{fig:edge_loss_evolution}
\end{figure}

In this 20-client setting, the number of AMGF collaboration groups is not fixed a priori, but re-estimated at each communication round by Affinity Propagation (AP) from the round-specific momentum-affinity matrix $\mathcal{A}^{t}$. The AP preference parameter was set to the median input similarity, so the cluster count $K^{t}$ was automatically determined by the client trajectory-similarity structure at each round.
Fig.~\ref{fig:edge_loss_evolution} compares the local RMSE evolution of six representative edges covering the main behavioral regimes: z54 represents zones well aligned with the shared building dynamics, z58 and z68 show the largest divergence between global and personalized learning, z27 and z69 are the personalization-demanding clients identified in \cite{HiFEL-OCKT}, and z41 exhibits the burstiest learning trajectory.

Fig.~\ref{fig:edge_loss_evolution} first shows that full-model capacity does not translate into better local accuracy under heterogeneous temporal dynamics. Trad-FL/FSL exhibits the highest initial RMSE on nearly all illustrated edges and remains dominated across most rounds on z58 and z68; at round $6$ of z68, the four configurations order exactly along the ablation chain, with RMSE of approximately $0.48$, $0.40$, $0.24$, and $0.18$ for Trad-FL/FSL, FSL-KD, AMGF without anticipation, and full AMGF, respectively. Overall, TeRR-SAtt performs at least as well as the full-capacity model on most edges and rounds in terms of learning accuracy while requiring a fraction of the edge-side resources (Figs.~\ref{fig:edge_train_inference_latency}--\ref{fig:edge_train_inference_performance}), confirming that its efficiency and autonomy gains do not come at the cost of learning effectiveness.

Overall, the results show that replacing the globally shared teacher update of FSL-KD with AMGF updates improves local learning behavior across heterogeneous clients.
Across the representative edges, full AMGF achieves maximum RMSE reductions of up to $35.31\%$, while AMGF without anticipation still reaches improvements of up to $31.49\%$ over FSL-KD.
This indicates that naive global aggregation can inject incompatible update directions into the teacher, whereas AMGF better preserves client-specific temporal learning requirements.
For edge z27, FSL-KD exhibits consistently higher and more unstable RMSE values than both AMGF variants. At round $4$, AMGF without anticipation reduces RMSE from $0.9116$ to $0.6246$, corresponding to a $31.49\%$ improvement, while full AMGF reaches a $25.90\%$ improvement. This suggests that the globally aggregated teacher update is influenced by heterogeneous client signals that do not fully align with the learning trajectory required by this edge. In contrast, AMGF reduces this effect by assigning the edge to a trajectory-compatible collaboration group and generating a cluster-specialized teacher update.

For edge z54, AMGF also outperforms FSL-KD across most rounds. The benefit is particularly visible around round $8$, where full AMGF reduces RMSE from $0.5311$ to $0.3672$, achieving a $30.86\%$ improvement over FSL-KD. This indicates that the adaptive look-ahead term helps exploit stable trajectory information when the cluster direction is sufficiently reliable. At the same time, AMGF without anticipation also reaches a $28.35\%$ improvement, confirming that path-aware momentum clustering and specialized teacher fusion improve federated split distillation, with anticipation providing further acceleration.

Edge z69 further highlights the benefit of momentum-guided trajectory tracking combined with anticipation. Full AMGF achieves the strongest observed gain, reducing RMSE from $0.7390$ to $0.4781$ at round $5$, i.e., a $35.31\%$ improvement over FSL-KD. This gain is larger than the $21.71\%$ improvement obtained without anticipation at the same round, showing that the anticipation term can substantially strengthen teacher adaptation when the trajectory direction is coherent. Thus, the gain on z69 confirms that AMGF groups compatible trajectories and effectively exploits their momentum to accelerate personalization under evolving non-I.I.D. temporal dynamics.
In the final rounds, the non-anticipatory variant is slightly better as the RMSE rises again, indicating that highly changing temporal dynamics may sometimes favor a more conservative update. Nevertheless, full AMGF remains beneficial over most of the trajectory, suggesting that anticipation is effective when the learned direction is sufficiently stable, while leaving room for further adaptive refinement.

Finally, edge z41 completes the analysis under a more bursty learning trajectory. Although all methods experience sharp RMSE variations, full AMGF reduces RMSE from $0.5807$ to $0.4150$ at round $4$, corresponding to a $28.53\%$ improvement over FSL-KD. This shows that adaptive anticipation can improve robustness when the client trajectory changes abruptly but remains directionally informative.
Overall, the ablation confirms the complementary roles of the two AMGF components. AMGF without anticipation validates the benefit of path-aware clustering and cluster-specialized teacher fusion, achieving up to $31.49\%$ RMSE reduction over global FSL-KD. Full AMGF further improves learning when trajectory directions are sufficiently coherent, reaching up to $35.31\%$ RMSE reduction and showing that reliability-gated anticipation can accelerate convergence and improve recovery under evolving non-I.I.D. temporal dynamics.

Across all 20 clients, with performance averaged over the 10 rounds, full AMGF achieves a mean RMSE reduction of $5.14\% \pm 6.14\%$ over FSL-KD. Client-level outcomes reach up to an $18.10\%$ RMSE reduction, with a $5.40\%$ increase observed in the least favorable case. AMGF without anticipation achieves a mean reduction of $0.51\% \pm 7.48\%$. These aggregate results complement the per-round analysis: momentum-affinity clustering and cluster-specialized fusion achieve substantial gains (up to $31.49\%$) on edges such as z27 and z69, which are identified in \cite{HiFEL-OCKT} as highly personalization-demanding clients whose learning objectives deviate from the shared dynamics of the LBNL building. Momentum-affinity clustering directly serves these demands by isolating such edges within trajectory-compatible collaboration groups. Reliability-gated anticipation then consolidates this specialization into a positive aggregate improvement across the heterogeneous client population. The two mechanisms are thus complementary: momentum-guided clustering discovers and serves client-specific personalization requirements, while anticipation stabilizes how far the resulting specialization is exploited.

\paragraph{Scalability Analysis.} The framework naturally extends to larger client populations. On the learning side, AMGF does not rely on a fixed global consensus: the number of collaboration groups $K^{t}$ is re-estimated at each round by AP, so a growing and increasingly heterogeneous client population is decomposed into additional, adaptively formed trajectory-compatible clusters rather than being forced into a single biased teacher update. On the deployment side, each edge device deploys its own TeRR-SAtt instance dimensioned to its local resource budget in Eq.~\eqref{eq:local-client-objective} and performs fully autonomous inference, so the measured edge-side efficiency is independent of the number of participating clients. The remaining cost is confined to the server: the momentum update in Eq.~\eqref{eq:momentum-equation} and teacher fusion scale linearly with $N$, whereas only the momentum affinity matrix construction and Affinity Propagation clustering scale quadratically.
This quadratic complexity arises from the clustering stage rather than from the fusion and anticipation mechanisms themselves, as Eqs.~\eqref{eq:cluster-gradient-momentum}--\eqref{eq:cluster-teacher-update} operate only on the resulting partition $\mathcal{G}^{t}$ and are therefore independent of the clustering algorithm used to produce it.

\section{Conclusion}
In this paper, we investigated personalized temporal edge intelligence under the joint constraints of edge efficiency, heterogeneous learning trajectories, and autonomous inference.
We proposed TeRR-SAtt to separate training guidance from edge deployment, allowing clients to benefit from collaborative teacher knowledge while retaining a lightweight local inference path.
We introduced AMGF to replace globally shared teacher updates with learning-momentum fusion for specialized learning guidance through clustering and adaptive anticipation.
Experiments on real-world smart-building data show that TeRR-SAtt improves the efficiency--autonomy trade-off, reducing edge training latency by up to $65.50\%$, inference latency by $44.70\%$, training memory usage by $18.40\%$, and inference CPU usage by $33.10\%$, while avoiding the server dependency inherent to FSL inference.
In addition, AMGF improves local learning performance over global updates, achieving up to $35.31\%$ RMSE reduction with anticipation and up to $31.49\%$ without anticipation, confirming the complementary benefits of path-aware clustering and reliability-gated anticipation.

Future work will extend the evaluation to additional smart-building and industrial IoT datasets and compare AMGF against personalized and clustered federated learning baselines.
We intend to investigate hierarchical momentum modeling across heterogeneous edge ecosystems, where client trajectories may evolve at multiple spatial and organizational levels.
Moreover, we plan to exploit momentum information beyond personalization, using it to reduce the communication overhead naturally associated with split learning.

\bibliographystyle{unsrt}
\bibliography{bib}

\appendix
\renewcommand{\theHsection}{appendix.\Alph{section}}

\section{Geometric Intuition: Anticipation and Personalized Collaboration}
\label{app:amgf-geometric-intuition}

\begin{figure}[htb]
    \centering
    \includegraphics[width=\linewidth]{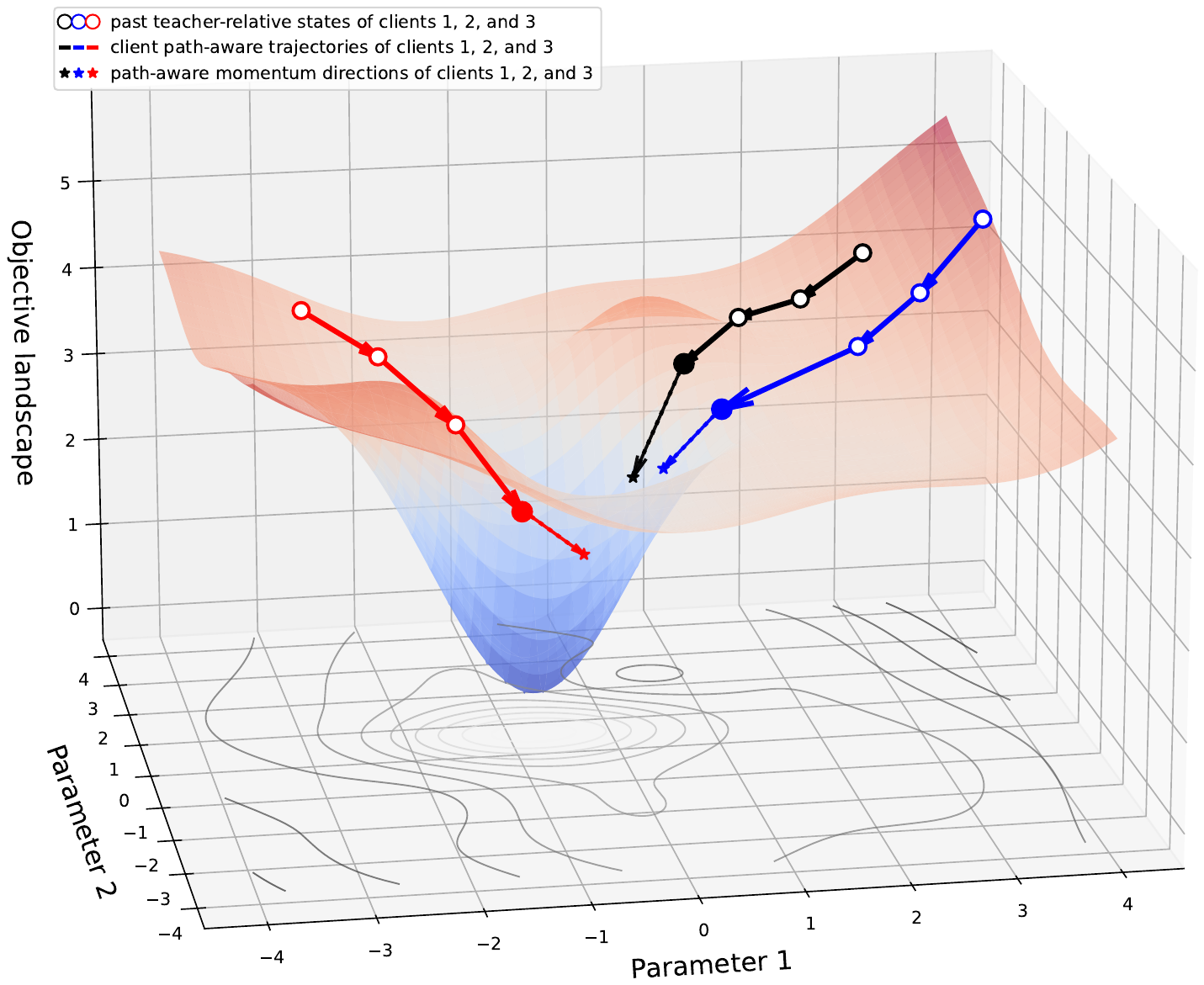}
    \caption{Geometric intuition of AMGF. The recent teacher-relative trajectories of three clients are shown on an illustrative objective landscape. Clients $1$ and $2$ exhibit similar path-aware trajectories and aligned momentum directions, making them suitable for personalized collaboration through a shared cluster-specialized teacher update. Client $3$ follows a distinct trajectory, motivating a separate collaboration group. The path-aware momentum directions summarize the persistent teacher-update tendencies used by AMGF for trajectory-aware clustering and reliable anticipation.}
    \label{fig:amgf-geometric-intuition}
\vspace{-5mm}
\end{figure}

Fig.~\ref{fig:amgf-geometric-intuition} provides a geometric intuition for the role of path-aware momentum in AMGF.
Each curve represents the recent teacher-relative trajectory experienced by one client across previous rounds.
The final arrow associated with each trajectory corresponds to the client's path-aware momentum direction $\mathbf{m}_{i}^{t}$, which summarizes the persistent teacher-displacement tendency induced by that client.
The key observation is that clients may be compatible even if their instantaneous gradients are not identical.
What matters for AMGF is whether their recent teacher-relative trajectories indicate similar displacement tendencies.
In the figure, clients $1$ and $2$ follow visually similar paths and induce aligned momentum directions.
They are, therefore, natural candidates for the same collaboration group, since they are likely to benefit from a shared cluster-specialized teacher update.
In contrast, client $3$ follows a different trajectory and induces a distinct momentum direction, suggesting that forcing it to share the same teacher update could dilute personalization or introduce conflicting adaptation signals.

This illustrates the motivation behind clustering clients using path-aware momentum.
Rather than grouping clients only according to current gradients or static data similarity, AMGF clusters clients according to the evolution of their learning trajectories in the teacher-parameter space.
As a result, the resulting collaboration groups reflect compatible update histories and support personalized teacher specialization.

The same geometric view also motivates AMGF's anticipation mechanism.
Once a cluster contains clients with coherent path-aware momentum directions, the cluster momentum $\mathbf{m}_{[k]}^{t}$ provides a reliable direction along which the teacher can be moved.
AMGF then applies a conservative cluster update and extends it through the anticipation coefficient $\alpha_{[k]}^t$ only when the cluster also exhibits strong intra-cluster agreement and temporal consistency.
Thus, anticipation is not applied as a fixed extrapolation step; it is activated only when the geometry of the client trajectories indicates that the cluster direction is sufficiently reliable.

\section{Behavioral Analysis of AMGF Coefficients}
\label{app:amgf-behavior-appendix}

This appendix provides additional details and rigorous analysis of the control quantities used by AMGF.
The main paper defines the path-aware anticipated teacher update in Eq.~\ref{eq:cluster-teacher-update}; here, we formalize how the intra-cluster agreement $A_{[k]}^t$ and the temporal trajectory stability $C_{[k]}^t$ mathematically regulate the look-ahead optimization space for learning anticipation.

\begin{table}[htbp]
\vspace{-5mm}
    \centering
    \caption{Functional roles of AMGF control quantities.}
    \label{tab:amgf_quantities}
    \small
    \renewcommand{\arraystretch}{1.3}
    \begin{tabular}{c p{2cm} p{8.0cm}}
        \hline
        \textbf{Quantity} & \textbf{Definition} & \textbf{Optimization Role in AMGF} \\
        \hline
        $A_{[k]}^{t}$ 
        & Eq.~\ref{eq:cluster-agreement}
        & Measures the pairwise directional agreement among the path-aware momentum vectors of clients assigned to cluster $G_k^t$. It quantifies whether the cluster represents a coherent teacher-displacement direction rather than a mixture of conflicting update requirements. \\
        
        $C_{[k]}^{t}$ 
        & Eq.~\ref{eq:trajectory-stability}
        & Measures the temporal consistency of client momentum directions across consecutive rounds. It indicates whether the cluster's anticipated direction is supported by stable trajectory evolution rather than by a transient or noisy displacement. \\
        \hline
    \end{tabular}
\vspace{-5mm}
\end{table}

The role of these quantities is to prevent AMGF from extrapolating unreliable optimization directions. Since the teacher update is expressed relative to the initial teacher state $\boldsymbol{\theta}_{\mathcal{T}}^{0}$, the cluster momentum $\mathbf{m}_{[k]}^{t}$ summarizes a smoothed teacher-displacement request for cluster $G_k^t$. However, using this displacement for look-ahead specialization is only desirable when two conditions hold simultaneously: the clients in the cluster should agree on the direction of the displacement, and this direction should remain temporally stable.
To enforce this condition, we define the adaptive anticipation coefficient in Eq.~\ref{eq:alpha-equation}, where $\alpha_{\max}$ controls the maximum admissible look-ahead magnitude. Since both $A_{[k]}^t$ and $C_{[k]}^t$ are bounded in $[0,1]$, the anticipation factor also satisfies:
\begin{equation}
    0 \leq \alpha_{[k]}^t \leq \alpha_{\max}.
    \label{eq:alpha-boundaries}
\end{equation}
Thus, $\alpha_{\max}$ defines the largest possible anticipation horizon, while the product $A_{[k]}^t C_{[k]}^t$ acts as a reliability gate.
This multiplicative form is intentionally conservative. If the cluster contains clients with incompatible path-aware momentum directions, then $A_{[k]}^t$ becomes small and anticipation is suppressed. Similarly, if individual client trajectories are unstable across rounds, then $C_{[k]}^t$ decreases, and the look-ahead term is reduced. Therefore, AMGF activates strong anticipation only when the cluster exhibits both spatial coherence and temporal stability.

\begin{table}[htbp]
\vspace{-5mm}
    \centering
    \caption{Interpretation of AMGF behavior under limiting values of $A_{[k]}^{t}$ and $C_{[k]}^{t}$.}
    \label{tab:amgf_cases}
    \small
    \renewcommand{\arraystretch}{1.35}
    \begin{tabular}{c c c c p{6cm}}
        \toprule
        \textbf{Case} 
        & $\boldsymbol{A_{[k]}^{t}}$ 
        & $\boldsymbol{C_{[k]}^{t}}$ 
        & \textbf{Operational Regime} 
        & \textbf{Interpretation} \\
        \midrule            
        1 
        & $\approx$$0$ 
        & $\approx$$0$ 
        & $\alpha_{[k]}^t \approx 0$.
        & The cluster lacks both internal directional agreement and temporal consistency in terms of path-aware momentum. The teacher updates fall back to the conservative momentum vector with no anticipation. \\[4pt]
        
        2 
        & $\approx$$0$ 
        & $\approx$$1$ 
        & $\alpha_{[k]}^t \approx 0$.
        & Clients may follow stable individual trajectories, but their directions are mutually incompatible. Look-ahead specialization is avoided to prevent conflicting teacher displacement. \\[4pt]
        
        3 
        & $\approx$$1$ 
        & $\approx$$0$ 
        & $\alpha_{[k]}^t \approx 0$.
        & Clients currently agree within the cluster, but their trajectories are not stable across rounds. AMGF avoids extrapolating a potentially transient direction. \\[4pt]
        
        4 
        & $\approx$$1$ 
        & $\approx$$1$ 
        & $\alpha_{[k]}^t \approx \alpha_{\max}$.
        & Clients exhibit both strong cluster-level agreement and stable trajectory evolution. AMGF extends the teacher update along the cluster momentum direction. \\
        \bottomrule
    \end{tabular}
\vspace{-5mm}
\end{table}

Under the AMGF update rule, the cluster-specialized teacher is given in Eq.~\ref{eq:cluster-teacher-update}.
When $\alpha_{[k]}^t \rightarrow 0$, the update reduces to the conservative cluster momentum vector:
\begin{equation}
    \boldsymbol{\theta}_{\mathcal{T},[k]}^{t+1}
    =
    \boldsymbol{\theta}_{\mathcal{T}}^{0} + \eta \mathbf{m}_{[k]}^{t}.
\end{equation}
In contrast, when $A_{[k]}^t \rightarrow 1$ and $C_{[k]}^t \rightarrow 1$, the update approaches the maximum look-ahead regime:
\begin{equation}
    \boldsymbol{\theta}_{\mathcal{T},[k]}^{t+1}
    =
    \boldsymbol{\theta}_{\mathcal{T}}^{0} + \bigl(\eta + \alpha_{\max}\bigr)\mathbf{m}_{[k]}^{t}.
\end{equation}
Hence, AMGF does not rely on a fixed anticipation step. Instead, it adapts the effective teacher displacement based on the reliability of the cluster trajectory.

This analysis shows that AMGF treats learning anticipation as a reliability-controlled extrapolation mechanism. The cluster momentum $\mathbf{m}_{[k]}^{t}$ determines the direction of teacher specialization, while $A_{[k]}^t$ and $C_{[k]}^t$ determine how far this direction can be safely extrapolated. Consequently, pairwise momentum agreement is not only used for clustering but also reappears as a post-clustering reliability control that prevents over-specialization when the resulting group is internally inconsistent.

\section{Stability and Bounded-Drift Properties of the Anchored Teacher Update}
\label{app:stability-bounded-drift}
We express each cluster-specialized teacher as a displacement from the initial teacher $\boldsymbol{\theta}_{\mathcal{T}}^{0}$ rather than as an unconstrained recursive update from the previous cluster teacher. This design is intentional. Let the cluster momentum be bounded as $\parallel \mathbf{m}_{[k]}^{t} \parallel \leq \mathbf{M}$ and let the adaptive anticipation factor satisfy $0 \leq \alpha^t_{[k]} \leq \alpha_{\max}$. From Eq.~\ref{eq:cluster-teacher-update}, we obtain:
\begin{equation}
    \parallel \boldsymbol{\theta}_{\mathcal{T},[k]}^{t+1} - \boldsymbol{\theta}_{\mathcal{T}}^{0} \parallel
    =
    \parallel (\eta + \alpha^t_{[k]}) \mathbf{m}_{[k]}^{t} \parallel
    \leq
    (\eta+\alpha_{\max}) \mathbf{M}.
\end{equation}
Thus, the teacher displacement remains bounded whenever the path-aware cluster momentum is bounded. Since $\mathbf{m}_{[k]}^{t}$ is obtained from exponentially smoothed client displacement requests, and the anticipation coefficient is gated by intra-cluster agreement and temporal consistency, we prevent repeated look-ahead steps from accumulating as an unconstrained drift.

In contrast, a recursive update of the form
\begin{equation}
    \boldsymbol{\theta}_{\mathcal{T},[k]}^{t+1} = \boldsymbol{\theta}_{\mathcal{T},[k]}^{t} + \left( \eta+\alpha^t_{[k]} \right) \mathbf{m}_{[k]}^{t}
\end{equation}
would accumulate all previous cluster-specialized displacements. Under changing cluster assignments, this may propagate stale or cluster-incompatible teacher directions across rounds, especially when clients move between collaboration groups. Anchoring the update to $\boldsymbol{\theta}_{\mathcal{T}}^{0}$ avoids this accumulation by interpreting $\mathbf{m}_{[k]}^{t}$ as the current smoothed displacement request relative to a common reference teacher. Therefore, all cluster-specialized teachers remain comparable within the same teacher-parameter coordinate system.

\end{document}